\documentclass[10pt]{article}

\usepackage{apacite}
\usepackage{cogsci}
\usepackage{pslatex}

\usepackage{amsmath}
\usepackage{caption}
\usepackage{xcolor}
\usepackage{graphicx}
\usepackage{multirow}
\usepackage{pgfplots}
\usepackage{subcaption}
\usepackage{tabularx}
\usepackage{tikz}
\usepackage{times}
\usepackage{url}
\usepackage{xspace}

\usetikzlibrary{positioning}
\pgfplotsset{compat=1.5,
    /pgfplots/ybar legend/.style={
        /pgfplots/legend image code/.code={%
        \draw[##1,/tikz/.cd,bar width=3pt,yshift=-0.2em,bar shift=0pt]
                plot coordinates {(0cm,0.8em)};},
    },
    every non boxed x axis/.append style={x axis line style=-},
     every non boxed y axis/.append style={y axis line style=-}
}

\newcommand\ie{\emph{i.e.}}
\newcommand\eg{\emph{e.g.}}

\newcommand{\ignore}[1]{}

\newcommand{\Figure}[1]{Figure~\ref{#1}}

\newcommand\XT{XT07\xspace}
\newcommand\SPSS{SPSS11\xspace}
\newcommand\NGS{NGS15\xspace}

\newcommand\subgroup{_{\, \mathrel{\raisebox{-1pt}{\scriptsize$\mathcal{G}$}}}}

\definecolor{sub}{HTML}{4D4D4D} 
\definecolor{basic}{HTML}{5DA5DA} 
\definecolor{sup}{HTML}{FAA43A}

\title{The Interaction of Memory and Attention in Novel Word Generalization: \\ A Computational Investigation}

\author{
    {\large \bf Erin Grant}, 
    {\large \bf Aida Nematzadeh}, 
    and 
    {\large \bf Suzanne Stevenson} 
    \\
    Department of Computer Science \\
    University of Toronto \\
    \{eringrant, aida, suzanne\}@cs.toronto.edu
}

\begin{document}

\maketitle

\begin{abstract}

People exhibit a tendency to generalize a novel noun to the basic-level in a hierarchical taxonomy -- a cognitively salient category such as ``dog'' -- with the degree of generalization depending on the number and type of exemplars.
Recently, a change in the presentation timing of exemplars has also been shown to have an effect, surprisingly reversing the prior observed pattern of basic-level generalization.
We explore the precise mechanisms that could lead to such behavior by extending a computational model of word learning and word generalization to integrate cognitive processes of memory and attention.
Our results show that the interaction of forgetting and attention to novelty, as well as sensitivity to both type and token frequencies of exemplars, enables the model to replicate the empirical results from different presentation timings.
Our results reinforce the need to incorporate general cognitive processes within word learning models to better understand the range of observed behaviors in vocabulary acquisition.


\end{abstract}

\section{Introduction}

A number of computational models have successfully mimicked child behaviors in learning the meaning of words from ambiguous input \shortcite<e.g.,>{siskind.1996,yu.ballard.2007,frank.etal.2007, fazly.etal.2010.csj}.
However, one challenge in word-meaning acquisition that has received less attention is that of \textit{novel word generalization}:  i.e., correctly identifying the level of a hierarchical taxonomy that a word refers to. After hearing it only a few times, how does the child determine, for example, that the word \textit{dog} refers to Dalmatians, all dogs of different breeds, or any kind of animal? This issue poses difficulties to the learner because the accumulated evidence can be compatible with more than one of these choices.
In this example, all dogs are also animals, and thus the meaning ``animal'' might also be consistent with all the usages of the word \textit{dog}. 

\citeA{xu.tenenbaum.2007.psyrev} (henceforth \XT) studied novel word generalization in both children and adults by observing decisions about category membership for novel objects in various experimental settings.
One of their important findings concerned how people responded having seen 1 vs.\ 3 labeled exemplars of a certain kind of entity within a taxonomy.
For example, having seen a single Dalmatian labeled as a \textit{fep}, people assumed that the novel word \textit{fep} could refer to the general category of dogs.
However, if people saw several Dalmatians called \textit{fep}, they apparently recognized that it would be a \textit{suspicious coincidence} if \textit{fep} meant ``dog'', but only one breed of dog was observed.
In such cases, people  had a lesser tendency to generalize to the higher level category than after seeing a single exemplar.

\citeA{spencer.etal.2011} (henceforth \SPSS) investigated the effect of presentation timing in the same task.
They found that presenting exemplars in sequence -- as opposed to simultaneously, as in \XT -- reverses the suspicious coincidence effect.
That is, after sequentially viewing three exemplars consistent with a more specific level of the taxonomy (e.g., three dogs of a single breed), people have a \textbf{greater} tendency to generalize to the higher category than after seeing one exemplar. 
\SPSS explained this reversal as an interaction of word learning with the more general cognitive processes of attention and memory, which differ in their operation across the presentation types:
People attend to and remember finer-grained similarities among objects when viewed simultaneously (e.g., that they are all Dalmatians), while the sequential presentation leads to  a consideration of the objects that focuses on their general commonalities (e.g., that they are all dogs).

Our goal in this paper is to provide a computational model that accounts for both the \XT and \SPSS findings in a well-motivated manner, by incorporating memory and attentional constraints into an incremental model of word learning and word generalization.
It is desirable to integrate together all these pieces -- novel word generalization, incremental word learning, and memory and attention -- because: (i) word generalization is part and parcel of learning the meaning of words, since it allows the abstraction of meaning from a sequence of specific experiences, and (ii) many word-learning behaviors are influenced by the general cognitive processes of memory and attention \shortcite<\eg,>{vlach.etal.2008,samuelson.smith.2000}.
Importantly, by explicitly specifying such mechanisms in a computational model, we contribute to the precise understanding of the interactions between them that are required to account for empirical data.

\section{Suspicious Coincidence: Data and Models}

\XT and \SPSS explored how people generalized a novel word like \textit{fep} to various levels of a taxonomy of objects (including animals, vehicles, and vegetables).
Basic-level categories are those whose members share a significant number of salient attributes (e.g., dogs or trucks);
subordinate categories occur lower in the hierarchy, and their members share many fine-grained attributes (e.g., Dalmatians or bulldozers);
superordinate categories (e.g., animals and vehicles) are higher than the basic-level, and their members have fewer attributes in common \cite{rosch.1973}.

For the sake of space, we focus only on two of the training conditions in \XT and \SPSS -- the ``1-example'' and ``3-subordinate'' conditions -- in which the suspicious coincidence effect and its reversal are seen.\footnotemark
\footnotetext{Our model replicates the results of \XT and \SPSS for all training conditions, but we only report the results for these two here.}
The 1-example condition has one training trial in which participants observe a single object (\eg, a Dalmatian) that is labeled with a novel word (such as \textit{fep}).
In the 3-subordinate condition, participants observe three instances from the same subordinate category (\eg, three different Dalmatians) labeled with the novel word.
In \XT's experiment, all three instances were presented  \textit{simultaneously}.
\SPSS included a condition in which the three instances are shown and labeled \textit{sequentially}.
(Simultaneous and sequential are the same for one example.)

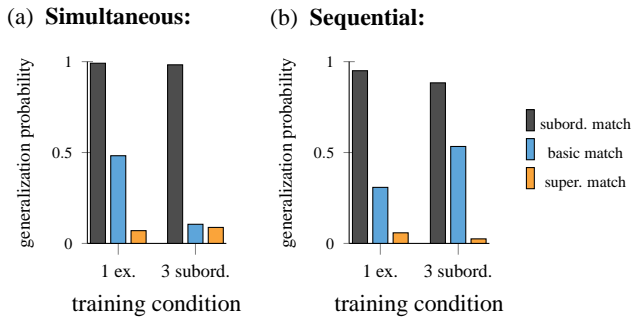
\begin{figure}[!h]

\caption{\textbf{\SPSS Behavioural Data}}

\captionsetup[subfigure]{justification=justified,singlelinecheck=false}
\begin{subfigure}[b]{0.4\linewidth}
\centering

\caption{\textbf{
    Simultaneous:
}}

\begin{tikzpicture}
\begin{axis}[
    axis lines=left,
    ybar,
    xlabel={training condition},
    xlabel style={font=\footnotesize},
    ylabel style={font=\scriptsize},
    ylabel={generalization probability},
    ytick={0, .5, 1},
    ymin=0, ymax=1,
    symbolic x coords={1 ex.,3 subord.},
    xtick=data,
    xticklabel style={font=\scriptsize},
    yticklabel style={font=\tiny},
    height=4cm,width=\textwidth,
    enlarge x limits=0.4,
    bar width=0.2cm
    ]
    
    \addplot[fill=sub] table [x=condition, y=sub. match, col sep=comma]
    {spencer_experiment_1.txt};
        
    \addplot[fill=basic] table [x=condition, y=basic match, col sep=comma]
    {spencer_experiment_1.txt};
        
    \addplot[fill=sup] table [x=condition, y=super. match, col sep=comma]
    {spencer_experiment_1.txt};
\end{axis}
\end{tikzpicture}
\label{fig:spencer-sim}
\end{subfigure}
\begin{subfigure}[b]{0.6\linewidth}
\centering
\caption{\textbf{
    Sequential:
}}
\begin{tikzpicture}
\begin{axis}[
    axis lines=left,
    ybar,
    legend style={draw=none,font=\tiny},
    legend style={at={(1.2,0.5)},anchor=west},
    xlabel={training condition},
    xlabel style={font=\footnotesize},
    ylabel style={font=\scriptsize},
    ylabel={generalization probability},
    ytick={0, .5, 1},
    ymin=0, ymax=1,
    symbolic x coords={1 ex.,3 subord.},
    xtick=data,
    xticklabel style={font=\scriptsize},
    yticklabel style={font=\tiny},
    height=4cm,width=0.67\textwidth,
    enlarge x limits=0.4,
    bar width=0.2cm
    ]
    
    \addplot[fill=sub] table [x=condition, y=sub. match, col sep=comma]
    {spencer_experiment_2.txt};
    \addlegendentry{subord. match}
        
    \addplot[fill=basic] table [x=condition, y=basic match, col sep=comma]
    {spencer_experiment_2.txt};
    \addlegendentry{basic match}
        
    \addplot[fill=sup] table [x=condition, y=super. match, col sep=comma]
    {spencer_experiment_2.txt};
    \addlegendentry{super. match}
\end{axis}
\end{tikzpicture}
\label{fig:spencer-seq}
\end{subfigure}

\vspace{0.1cm}

\small{
    \SPSS data for 
    (\ref{fig:spencer-sim}) simultaneous and
    (\ref{fig:spencer-seq}) sequential presentations.
    Each bar is the percent of chosen test objects of each type of
    match:
    subord(inate), basic(-level), or super(ordinate).
    Differences in 1-ex.\ across the two experiments were not statistically significant.
}
\label{fig:spencer-data}
\vspace{-.4cm}
\end{figure}

After training, participants select all and only objects that they think are \textit{feps} from a set of test items.
Each test object is assessed as exactly one of the following types of match: 
\begin{itemize}
\vspace{-0.1cm}
\item a \textbf{subordinate match} has the same subordinate category as a training object (\eg, a Dalmatian).
\vspace{-0.2cm}
\item a \textbf{basic-level match} has the same basic-level category as a training object (\eg, a dog, but not a Dalmatian).
\vspace{-0.2cm}
\item a \textbf{superordinate match} has the same superordinate category as training objects (\eg, an animal other than a dog).
\vspace{-0.4cm}
\end{itemize}
Since \SPSS replicated the pattern found by \XT in their simultaneous presentation condition, we report only the results of \SPSS, as shown in Fig.~\ref{fig:spencer-data}.

In the 1-example condition people generalized the novel word to refer to both subordinate matches (\eg, Dalmatians) and (to a lesser extent) basic-level matches (\eg, other kinds of dogs), but not to the superordinate matches (\eg, other animals).
This is in line with the idea that people tend to generalize a novel word to a basic-level category such as ``dog'' because of the perceptual salience of this level of categorization \cite<\eg,>{markman.1991}.%,golinkoff.etal.1994}.

In the 3-subordinate condition, when objects are presented simultaneously (\Figure{fig:spencer-sim}), the generalization to the basic level is attenuated compared to the 1-example condition.
\XT explained this behavior as the suspicious coincidence effect.
However, when objects are presented sequentially (\Figure{fig:spencer-seq}), there was a surprising reversal of this effect.

While \SPSS outline possible memory and attentional processes to explain their results,  we know of no computational model that can account for both sets of data.
\XT's Bayesian model formed hypotheses over a detailed hierarchical taxonomy to account for their own data, but it cannot model the difference between presentation timings, as \SPSS note.
The computational word learner of \citeA{nematzadeh.etal.2015.emnlp} (henceforth \NGS) can model the \XT results without the need for elaborated knowledge of the hierarchy or a built-in basic-level bias.
Instead, the results of the model arise from a general type-token frequency interaction of the sort that commonly arises in explanations of linguistic phenomena \cite<e.g.,>{bybee.1985,croft.cruse.2004}.
However, the timing of presentations also has no effect on the \NGS model, and so the reversal of the suspicious coincidence effect is not achieved.
In the next section, we explain how the \NGS model can be naturally extended to integrate memory and attention, and therefore sensitivity to presentation timing.

\section{Our Computational Model}

We start with the \NGS model because it uses  an incremental word learning framework that mimics a range of behaviors in vocabulary acquisition \shortcite<\eg,>{fazly.etal.2010.csj,fazly.etal.2010.cogsci}. %, nematzadeh.etal.2012.cogsci}.
This framework has recently been extended to incorporate the effects of memory and attention on word learning \cite{nematzadeh.etal.2012.cmcl}, presenting a natural opportunity for integrating these processes within word generalization.
We describe the \NGS model, then the novel extensions that enable our model to replicate the \SPSS data.

\subsection{Learned Meanings in the \NGS Model}

The \NGS model is a cross-situational learner that tracks weighted co-occurrences of words and semantic features across its input as in \citeA{fazly.etal.2010.csj}.
The input to the model is intended to reflect the naturalistic input a child is exposed to, which consists of linguistic input (the words a child hears) paired with nonlinguistic data (the things a child perceives).
An input pair is the set of words $U_t$ and the set of semantic features $S_t$ observed at time $t$:
\begin{table}[h!]
\small{
\begin{tabular}{l}
{\bf $U_t$:} $\{$\, \emph{look},\; \emph{a},\; \emph{fep} $\}$ \\
{\bf $S_t$:} $\{$\, \textsc{perception},\, \textsc{look},\, \dots,\, \textsc{Dalmatian},\, \textsc{dog},\, \textsc{animal} $\}$
\end{tabular}
}
\end{table}

\noindent
The output of the model at each time $t$ is a set of meaning probabilities, $P_t(f_i|w_j)$, for each feature $f_i$ and each word $w_j$ observed up through time $t$.
The set of all conditional probabilities $P_t(f_i|w_j)$ for $w_j$ represents the meaning of $w_j$.

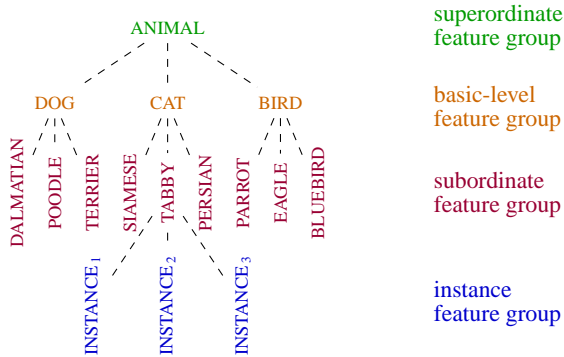
\begin{figure}[!h]
\centering
\scalebox{1}{
\begin{tikzpicture}[]

    \tikzstyle{level 1}=[level distance=1cm, sibling distance=1.5cm, font=\scriptsize]l
    \tikzstyle{level 2}=[level distance=1.2cm, sibling distance=0.5cm, font=\scriptsize]
    \tikzstyle{level 3}=[level distance=1.5cm, sibling distance=1cm, font=\scriptsize]

    \node[level 1] (superordinate) {\color{black!40!green}ANIMAL}
    child[dashed] {node[level 1] {\color{black!20!orange}DOG}
            child[dashed] {node[level 2] {\color{black!20!purple}\rotatebox{90}{DALMATIAN}}}
            child[dashed] {node[level 2] {\color{black!20!purple}\rotatebox{90}{POODLE}}}
            child[dashed] {node[level 2] {\color{black!20!purple}\rotatebox{90}{TERRIER}}}
        }
        child[dashed] {node[level 1] {\color{black!20!orange}CAT}
            child[dashed] {node[level 2] {\color{black!20!purple}\rotatebox{90}{SIAMESE}}}
            child[dashed] {node[level 2] {\color{black!20!purple}\rotatebox{90}{TABBY}}
                child[dashed] {node[level 3] {\color{black!20!blue}\rotatebox{90}{INSTANCE$\,_1$}}}
                child[dashed] {node[level 3] {\color{black!20!blue}\rotatebox{90}{INSTANCE$\,_2$}}}
                child[dashed] {node[level 3] {\color{black!20!blue}\rotatebox{90}{INSTANCE$\,_3$}}}
            }
            child[dashed] {node[level 2] {\color{black!20!purple}\rotatebox{90}{PERSIAN}}}
        }
        child[dashed] {node[level 1] {\color{black!20!orange}BIRD}
            child[dashed] {node[level 2] {\color{black!20!purple}\rotatebox{90}{PARROT}}}
            child[dashed] {node[level 2] {\color{black!20!purple}\rotatebox{90}{EAGLE}}}
            child[dashed] {node[level 2] {\color{black!20!purple}\rotatebox{90}{BLUEBIRD}}}
        };
        
        \node (sup label) [right=2.8cm of superordinate] {\footnotesize\parbox{2cm}{\color{black!40!green}superordinate\\feature group}};

        \node (basic level) [below=0.23cm of sup label] {\footnotesize\parbox{2cm}{\color{black!20!orange}basic-level\\feature group}};
        
        \node (sub label) [below=0.3cm of basic level] {\footnotesize\parbox{2cm}{\color{black!20!purple}subordinate\\feature group}};
        
        \node (inst label) [below=0.6cm of sub label] {\footnotesize\parbox{2cm}{\color{black!20!blue}instance\\feature group}};
        
\end{tikzpicture}
\vspace{-0.2cm}
}\caption{\small{
    A portion of the taxonomy used in this paper.
}}
\label{fig:tax}
\vspace{-0.4cm}
\end{figure}

The representation of meaning in \NGS reflects the structure of taxonomic knowledge.
Meaning features are arranged into \textit{feature groups}, each corresponding to a level of the taxonomic hierarchy, as shown in \Figure{fig:tax}.
For each word $w_j$, a meaning probability distribution, $P_t(.|w_j)$, is calculated for each feature group; that is, $P_t(.|w_j)$ is normalized over the features in a group, rather than over all meaning features.
The result is that features at the same level of the hierarchy, such as \textsc{Dalmatian} and \textsc{Poodle}, or \textsc{dog} and \textsc{cat}, compete for probability mass; this ensures that such features, which are mutually incompatible given their taxonomic relationship, cannot simultaneously have high probability.
Features at different levels of the hierarchy are in different feature groups and thus do not compete for probability mass; this ensures that meaning probabilities such as
$P_t(\textsc{Dalmatian}|\textit{fep})$, $P_t(\textsc{dog}|\textit{fep})$, and $P_t(\textsc{animal}|\textit{fep})$ can all be highly activated if \textit{fep} is intended to refer to a Dalmatian (which is also both a dog and an animal).  In this approach, the meaning of a word is the set of $n$ distributions, $P_t(.|w_j)$, one per feature group in a taxonomy with $n$ levels.

\subsection{The \NGS Learning Algorithm}

The input to the model is processed in an incremental two-step bootstrapping framework: Words and features that co-occur are aligned in proportion to the current meaning probabilities, which are then updated with the new evidence regarding strength of association, as follows.

\subsubsection{The Alignment Step.}

For an input pair at time $t$, the alignment strength for each feature $f_i \in S_t$ and word $w_j \in U_t$ is:
\begin{align}
    \label{eq:align}
    a_t(f_i, w_j) 
    &= \frac{P_{t-1}(f_i|w_j)}{\sum\limits_{w' \in U_t}{P_{t-1}(f_i|w')}} 
\end{align}

\noindent
These alignments are incrementally accumulated as:
\begin{align}
    \label{eq:assoc}
    \textrm{assoc}_{t}(f_i, w_j) 
    &= \sum_{t' \in \mathcal{T}} a_{t'} (f_i, w_j)
\end{align}
where $\mathcal{T}$ is all times at which $f_i$ and $w_j$ have co-occurred.

In our model, if more than one instance of a feature occurs at time $t$, multiple instances of the alignment for that feature and word are recorded.
For example, in the simultaneous presentation of three exemplars, $a_{t} (f_i, w_j)$ will contribute three times to the association score for each feature $f_i$ in the input.

\subsubsection{Update of Meaning Probabilities}

The model next uses the association scores to update the meaning probabilities.
Each meaning probability $P_{t}(f_i|w_j)$ represents the magnitude of the $f_i$--$w_j$ association \textit{relative to} the association strength between $w_j$ and other features within the same feature group $\mathcal{G}$ as $f_i$:
\begin{align}
    \label{eq:meaning-prob}
    P_{t}(f_i|w_j) 
    &= \displaystyle\frac{
        \textrm{assoc}_{t}(f_i, w_j) + 
        \gamma_{\, \mathrel{\raisebox{-1pt}{\scriptsize$\mathcal{G}$}}}^{\ t}
    }{
        \sum\limits_{f_{m} \in \mathcal{G}}\textrm{assoc}_{t}(f_m, w_j) + 
        k_{\mathrel{\raisebox{-1pt}{\scriptsize$\mathcal{G}$}}}
        \gamma_{\, \mathrel{\raisebox{-1pt}{\scriptsize$\mathcal{G}$}}}^{\ t}
    }
\end{align}
Here $k_{\mathrel{\raisebox{-1pt}{\scriptsize$\mathcal{G}$}}}$ and 
$\gamma_{\, \mathrel{\raisebox{-1pt}{\scriptsize$\mathcal{G}$}}}^{\ t}$
are smoothing terms: 
$k_{\mathrel{\raisebox{-1pt}{\scriptsize$\mathcal{G}$}}}$ 
reflects the expected number of features in $\mathcal{G}$
and
$\gamma_{\, \mathrel{\raisebox{-1pt}{\scriptsize$\mathcal{G}$}}}^{\ t}$ 
represents the \textit{a priori} tendency to observe a feature in $\mathcal{G}$.
While $k_{\mathrel{\raisebox{-1pt}{\scriptsize$\mathcal{G}$}}}$ is a fixed parameter, $\gamma_{\, \mathrel{\raisebox{-1pt}{\scriptsize$\mathcal{G}$}}}^{\ t}$ is a function of the number of observed types within the feature group $\mathcal{G}$, and thus changes over time (see \NGS).

The $\gamma\subgroup^{\ t}$ parameters are key to the generalization behavior of the \NGS model because they influence how much probability mass is allocated to a feature previously unseen with a word (cf.\ Eqn.~\ref{eq:meaning-prob} when the assoc score is $0$).
A higher value for $\gamma\subgroup^{\ t}$ leads to more probability mass allocated to previously unseen features in group $\mathcal{G}$, allowing for more generalization to new features in that group.
Because $\gamma\subgroup^{\ t}$ increases with the number of types, it captures the oft-observed tendency in language that people more readily generalize categories for which a greater variety of types of items has been observed.
The model matches the child data from \XT by equating $\gamma\subgroup^{\ 0}$ across feature groups. But to match the adults, who show a stronger basic-level bias, the model required that the $\gamma\subgroup^{\ 0}$ 
parameters be initialized to successively higher values for feature groups successively lower in the hierarchy, entailing that, e.g., it is easier to generalize a novel word to a new breed of dog not seen in training (basic-level generalization), than to a new kind of animal not seen in training (superordinate generalization).

We next describe how we incorporate mechanisms of memory and attention into the model, which render it sensitive to presentation timing.

\subsection{Our Extensions to Integrate Memory and Attention}

We adopt the general approach of \citeA{nematzadeh.etal.2012.cmcl} because it integrates memory and attention seamlessly into the cross-situational word-learning mechanism.
The approach was shown to account for spacing effects in word learning, which are closely related to the presentation timing factors considered by \SPSS.  
However, the methods must be extended to adequately meet the needs of word generalization in the \NGS model; we describe those extensions here.

\subsubsection{Modeling the Effects of Forgetting.}

To model the effect of memory, we use the association score formulation of \citeA{nematzadeh.etal.2012.cmcl}, which implements ``forgetting'' by applying a decay factor to each alignment probability (cf.~Eqn.~\ref{eq:assoc} above):
\begin{align}
    \label{eq:new-assoc}
    \textrm{assoc}_{t}(f_i, w_j) 
    &= \sum_{t' \in \mathcal{T}}
    %\left(
        \frac{
            a_{t'} (f_i, w_j)
        }{
            (t - t' + 1)^{d_{a_{t'}}}
        }
    %\right)
\end{align}
Each alignment in the sum is scaled by the temporal distance between the current time $t$ and the time $t'$ that the alignment was made, exponentiated to a decay function $d_{a_{t'}}$ that is inversely proportional to the strength of alignment.

However, we must extend this decay formulation to accommodate our hierarchical knowledge of feature groups.\footnotemark
\footnotetext{\citeA{nematzadeh.etal.2012.cmcl} used a single meaning probability distribution over all features -- i.e., there are no feature groups.}
In particular, we find that using the same decay rate across all feature groups is not sufficient.
As noted above, appropriate word generalization in the \NGS model requires that lower levels in the taxonomy be more ``open'' to generalizing to new features than higher levels in the taxonomy.
It is important to note that the decay of alignments also influences ``openness'' to generalization because it shifts probably mass away from observed word--feature pairs onto unseen events.
Thus, to appropriately reflect the nature of the hierarchy -- that openness increases with greater depth in the taxonomy -- we must parameterize decay by feature group.
Just as feature groups lower in the taxonomy must have successively higher $\gamma$ values to indicate more ``openness'' to generalization, lower feature groups also require higher decay rate parameters.

We thus use the following formulation of decay:
\begin{align}
    \label{eq:feature-group-decay}
    d_{a_{t}}
    &= \frac{
        d_{ \mathrel{\raisebox{-1pt}{\scriptsize$\mathcal{G}$}}}
    }{
        a_t(f_i, w_j)
    }
\end{align}
where $d_{ \mathrel{\raisebox{-1pt}{\scriptsize$\mathcal{G}$}}}$ controls the rate of decay for features in feature group $\mathcal{G}$, and is set successively higher for lower-level feature groups in the taxonomy.

\subsubsection{Modeling Attention to Novelty.}

Building on research showing that people attend more to novel stimuli in learning \shortcite<e.g.,>{snyder.etal.2008, macpherson.moore.2010, horst.etal.2011}, we use the general idea of  \citeA{nematzadeh.etal.2012.cmcl} in allocating more strength to alignments that are more novel  (cf.~Eqn.~\ref{eq:align}): 
\begin{align}
    \label{eq:new-align}
    a_t (f_i, w_j)
    &= \frac{P_{t-1}(f_i|w_j)}{\sum\limits_{w' \in U}{P_{t-1}(f_i|w')}} 
    \cdot
    \textrm{novelty}_t(f_i,w_j)
\end{align}
In this model, $\textrm{novelty}_t(f_i,w_j)$ was inversely proportional to how recently $w_j$ had been observed, and thus focused solely on novelty of words; the novelty of the feature $f_i$ was not considered.
We must broaden this approach because the experiments here are focused on a single novel word.

Here instead we consider the novelty of the observed \textit{word--feature pairing}, and again draw on considerations of type--token frequencies, as in other aspects of the \NGS model.
Specifically, we scale the alignment strength by the ratio of the token frequency of $f_i$--$w_j$ observations at time $t$ to the total frequency of all such observations, by formulating $\textrm{novelty}_t(f_i,w_j)$ as:
\begin{align}
    \label{eq:novelty}
    \textrm{novelty}_t (f_i, w_j)
    &= 
    \frac{
        \mathrm{token}_t\left(f_i, w_j\right)
    }{
        \sum_{t' \in \mathcal{T}}
        \mathrm{token}_{t'}\left(f_i, w_j\right)
    }
\end{align}
\noindent where $\mathrm{token}_t\left(f_i, w_j\right)$ is the number of tokens of feature $f_i$ that occurred at time $t$ with word $w_j$.

This formulation achieves attention to novelty as follows.
Generally, earlier observations of feature $f_i$ with word $w_j$ will have a stronger alignment than later observations, where the increased number of observations will increase the denominator of $\textrm{novelty}_t(f_i,w_j)$, and lead to attenuation of the alignment strength.
Note that when the co-occurrence of $f_i$ with word $w_j$ is truly novel -- \ie, the first time they are observed together -- the strength of alignment is undiminished, since the numerator and denominator of the novelty factor are equal in the initial observation of $f_i$ with $w_j$.

\subsubsection{Summary of Novel Extensions to the \NGS Model}

In summary, we have extended both the model of \NGS, and the memory and attention mechanisms of \citeA{nematzadeh.etal.2012.cmcl}, by: (i) incorporating a forgetting mechanism that is sensitive to the taxonomic level of a feature group, which reflects the needs of taxonomic structure and the process of novel word generalization; and (ii) formulating a mechanism for attention to novelty of word--feature pairings, rather than just to recency of words, consistent with the key role of word--feature association statistics in the model.

These mechanisms have a direct impact on the processing of stimuli in simultaneous vs.\ sequential presentations in a novel word generalization task.  The forgetting mechanism ensures that more general features, such as the kind of animal observed (\eg, dog or cat), are remembered better than more detailed features, such as particular breeds of dogs.  The attention-to-novelty mechanism has the consequence that successive observations of word--feature pairings in a sequential presentation scenario are ``discounted'' with respect to earlier presentations.
We demonstrate in our experiments below that, together, these mechanisms interact to enable the model to account for both the suspicious coincidence effect in a simultaneous presentation as found by \XT, and its reversal in a sequential presentation as found by \SPSS.

\section{Methodology}
\label{sec:method}

We follow the methods of \NGS, adapted where needed for our extended model on the \SPSS data.\footnotemark
\footnotetext{Our code and data are available at \url{https://github.com/eringrant/novel_word_generalization}.}

\vspace{.2cm}
\noindent\textbf{Training the Model.}  
We use a taxonomy with three levels, corresponding to the subordinate, basic, and superordinate categories of animals.
This yields four feature groups, one per category level plus an ``instance'' group to distinguish multiple objects of the same subordinate category.
See \Figure{fig:tax}.
In each $U_t$--$S_t$ input pair, $U_t$ consists of the novel word, and $S_t$ is a set of four features (one per feature group) representing a unique instance of the same subordinate category across all training trials; for example:
\begin{table}[!htbp]
\vspace{-.2cm}
\small{
\begin{tabular}{lrl}
%\multirow{2}{*}{1}
&{\bf $U_t$:} &$\{$\;\emph{fep}$\;\}$\\
&{\bf $S_t$:} &$\{$\;\textsc{instance$_1$}, \textsc{Dalmatian}, \textsc{dog}, \textsc{animal} $\}$\\\\
\end{tabular}
}
\label{tb:example-training}
\vspace{-0.5cm}
\end{table}

\noindent
In the 1-example condition, training consists of just one such $U_t$--$S_t$ pair.
In the 3-subordinate condition, training has three such $U_t$--$S_t$ pairs, differing only in the unique instance feature (\ie, instance$_1$, instance$_2$, instance$_3$) in each $S_t$.
In the simultaneous condition, the three $U_t$--$S_t$ pairs are all presented at the same time $t$.
In the sequential condition, the three $U_t$--$S_t$ pairs are presented one at a time, at $t$, $t+1$, and $t+2$.

\vspace{.2cm}
\noindent\textbf{Testing the Model.} After training, the level of generalization of the novel word is assessed against test objects, each of which is a subordinate match, a basic-level match, or a superordinate match; for example:
\begin{table}[!htbp]
\vspace{-0.1cm}
\small{
\begin{tabular}{lll}
{\bf subord. match:} &$\{$\;\textsc{instance$_4$}, \textsc{Dalmatian}, \textsc{dog}, \textsc{animal} $\}$\\\\
{\bf basic match:} &$\{$\;\textsc{instance$_5$}, \textsc{poodle}, \textsc{dog}, \textsc{animal} $\}$\\\\
{\bf super. match:} &$\{$\;\textsc{instance$_6$}, \textsc{toucan}, \textsc{bird}, \textsc{animal} $\}$\\\\
\end{tabular}
}
\label{tb:example-test}
\vspace{-0.5cm}
\end{table}

\noindent
We adapt the $P_{gen}$ formula of \NGS to test whether the model generalizes the learned meaning of the novel word $w$ to the various levels of match at test time $t$ (after training):
\begin{align*}
P_{gen}(m | w) = \frac{
\text{avg}_{Y\text{\tiny$\in$} m}\ P_t(Y|w) 
} {
\text{avg}_{Y'\text{\tiny$\in$ $\{$sub.$\}$}}\ P_t(Y'|w)
}
\end{align*}
\noindent
Here $P_t(Y|w)$ is the probability of a test object $Y$ given $w$, and
$m$ is the set of test objects at a certain level of match.
The measure in the numerator of $P_{gen}$ is the average such probability across test matches at that level, $\text{avg}_{Y\text{\tiny$\in$} m}\ P_t(Y|w)$.  This is not directly comparable to the empirical data, which are the percentages of test objects selected from each type of match.
To obtain a comparable measure, we scale each probability (for each level of match) by the probability of the subordinate matches in that condition,
$\text{avg}_{Y'\text{\tiny$\in$ $\{$sub.$\}$}}\ P_t(Y'|w)$ (the denominator of $P_{gen}$).
Thus $P_{gen}(m|w)$ is the relative average preference for test items at level $m$.
This renders the subordinate match probability as 1.0 (reflecting that people generally pick close to 100\% of the subordinate test items), and shows the other type of matches relative to that amount.

\vspace{.2cm}
\noindent\textbf{Model Parameters.}
\noindent
Since the \SPSS participants are adults, we use the \textit{adult parameter settings} of \NGS for the four
$\gamma\subgroup^{\ 0}$ parameters
and the four 
$k\subgroup$ (one each per feature group),
which are tuned to achieve a match to adult data of \XT.
For our decay parameters, we use:
\vspace{-0.2cm}
\begin{table}[h]
\small{
\centering
\bgroup
\setlength{\tabcolsep}{.5em}
\begin{tabular}{llll}
%\hline
$d_{\,\text{inst}} = 0.8$ & $d_{\,\text{subord}} = 0.5$ & 
$d_{\;\text{basic}} = 0.05$ & $d_{\,\text{super}} = 0.01$
%\hline
\end{tabular}
\egroup

%\vspace{-0.3cm}
\label{tb:adult-params}
%\vspace{-0.4cm}
}
\end{table}
\vspace{-0.4cm}

\section{Model Results and Discussion}

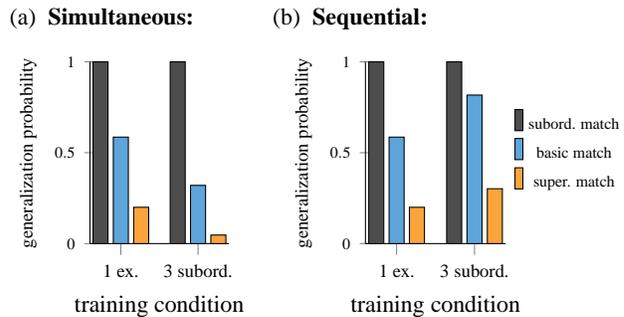
\begin{figure}[!h]

\caption{\textbf{Our Model Data}}

\captionsetup[subfigure]{justification=justified,singlelinecheck=false}
\begin{subfigure}[b]{0.4\linewidth}
\centering

\caption{\textbf{
    Simultaneous:
}}

\begin{tikzpicture}
\begin{axis}[
    axis lines=left,
    ybar,
    xlabel={training condition},
    xlabel style={font=\footnotesize},
    ylabel style={font=\scriptsize},
    ylabel={generalization probability},
    ytick={0, .5, 1},
    ymin=0, ymax=1,
    symbolic x coords={1 ex.,3 subord.},
    xtick=data,
    xticklabel style={font=\scriptsize},
    yticklabel style={font=\tiny},
    height=4cm,width=\textwidth,
    enlarge x limits=0.4,
    bar width=0.2cm
    ]
    
    \addplot[fill=sub] table [x=condition, y=sub. match, col sep=comma]
    {gammasup_0.2,gammabas_0.5,gammasub_1.0,gammainst_1.2,k_100.0,decaysup_0.01,decaybas_0.05,decaysub_0.5,decayinst_0.8,spacing_simultaneous,test_0.dat};
        
    \addplot[fill=basic] table [x=condition, y=basic match, col sep=comma]
    {gammasup_0.2,gammabas_0.5,gammasub_1.0,gammainst_1.2,k_100.0,decaysup_0.01,decaybas_0.05,decaysub_0.5,decayinst_0.8,spacing_simultaneous,test_0.dat};
        
    \addplot[fill=sup] table [x=condition, y=super. match, col sep=comma]
    {gammasup_0.2,gammabas_0.5,gammasub_1.0,gammainst_1.2,k_100.0,decaysup_0.01,decaybas_0.05,decaysub_0.5,decayinst_0.8,spacing_simultaneous,test_0.dat};
\end{axis}
\end{tikzpicture}
\label{fig:model-exp-1}
\end{subfigure}
\begin{subfigure}[b]{0.6\linewidth}
\centering
\caption{\textbf{
    Sequential:
}}
\begin{tikzpicture}
\begin{axis}[
    axis lines=left,
    ybar,
    legend style={draw=none,font=\tiny},
    legend style={at={(1,0.5)},anchor=west},
    xlabel={training condition},
    xlabel style={font=\footnotesize},
    ylabel style={font=\scriptsize},
    ylabel={generalization probability},
    ytick={0, .5, 1},
    ymin=0, ymax=1,
    symbolic x coords={1 ex.,3 subord.},
    xtick=data,
    xticklabel style={font=\scriptsize},
    yticklabel style={font=\tiny},
    height=4cm,width=0.67\textwidth,
    enlarge x limits=0.4,
    bar width=0.2cm
    ]
    
    \addplot[fill=sub] table [x=condition, y=sub. match, col sep=comma]
    {gammasup_0.2,gammabas_0.5,gammasub_1.0,gammainst_1.2,k_100.0,decaysup_0.01,decaybas_0.05,decaysub_0.5,decayinst_0.8,spacing_massed,test_0.dat};
    \addlegendentry{subord. match}
        
    \addplot[fill=basic] table [x=condition, y=basic match, col sep=comma]
    {gammasup_0.2,gammabas_0.5,gammasub_1.0,gammainst_1.2,k_100.0,decaysup_0.01,decaybas_0.05,decaysub_0.5,decayinst_0.8,spacing_massed,test_0.dat};
    \addlegendentry{basic match}
        
    \addplot[fill=sup] table [x=condition, y=super. match, col sep=comma]
    {gammasup_0.2,gammabas_0.5,gammasub_1.0,gammainst_1.2,k_100.0,decaysup_0.01,decaybas_0.05,decaysub_0.5,decayinst_0.8,spacing_massed,test_0.dat};
    \addlegendentry{super. match}
\end{axis}
\end{tikzpicture}
\label{fig:model-exp-2}
\end{subfigure}

\vspace{0.2cm}

\small{
    Our model data for 
    (\ref{fig:model-exp-1}) simultaneous and
    (\ref{fig:model-exp-2}) sequential presentations.
    Each bar is the probability of a type of test match: 
    \ie, subord(inate), basic(-level), or super(ordinate),
    scaled by the subordinate match probability.
}
\label{fig:model-exp}
\vspace{-0.5cm}
\end{figure}

\Figure{fig:model-exp} shows the results of our model in the simultaneous and sequential conditions; cf.~\Figure{fig:spencer-data} for the human behavioral data in \SPSS.
Following simultaneous presentation of training input (\Figure{fig:model-exp-1}), our model shows the suspicious coincidence effect: Generalization to the basic level is inhibited in the 3-subordinate condition as compared to the 1-example condition.
In contrast, sequential presentation reverses the suspicious coincidence effect (\Figure{fig:model-exp-2}): the model exhibits greater basic-level generalization in the 3-subordinate condition.
Thus, these results replicate the qualitative pattern evident in the behavioural data of \SPSS.

The interaction (between presentation type and amount of training) seen in the human data arises as a result of a corresponding interaction in the model.
Consider the comparison of each 3-subordinate condition (simultaneous and sequential) to the 1-example condition.
\textbf{In the simultaneous 3-subordinate case}, the attentional mechanism yields higher alignment strengths between the word and features because their three co-occurrences are all novel at the single presentation time; in addition there is little forgetting because the items are all seen at time $t$ and test is at time $t+1$.
This yields stronger subordinate alignments compared to the 1-example case, and therefore somewhat less basic-level generalization.

By contrast, \textbf{in the sequential 3-subordinate case}, the word--feature co-occurrences are less salient because they decrease in novelty over the three presentation times.
In addition, greater forgetting occurs because there is more time between the (first two) presentation times and test time ($t+4$).
In this case, because subordinate features decay faster than basic features, the interaction yields weaker subordinate features compared to the 1-example case, and more basic-level generalization is achieved.

The interaction of memory and attention effects are required to obtain this pattern of results in the model.
If the model includes only the decay mechanism, differentiated by taxonomic level, this enables it to focus more on abstract than specific features, and consequently raises the basic generalization closer to the level of the subordinate generalization in \textit{all} conditions.
On the other hand, using the attention mechanism alone enables the model to distinguish the sequential and simultaneous conditions, but it cannot on its own raise the basic generalization high enough.
Only when the two are used together does the model produce the reversal of the suspicious coincidence effect in the sequential presentation.

The necessity of both memory and attention is suggestive of how word learning occurs in people.
In particular, the attention mechanism in the model focuses more probability onto word--feature co-occurrences in their earlier presentations, simulating the general tendency for people to attend more to less-familiar things.
In addition, the mechanism that increases the decay rate for lower-level features in the taxonomy simulates the tendency in people to remember abstract features of objects over very specific features.
\SPSS contended that people are able to attend to specific features in the simultaneous condition due to the close spatial and temporal proximity of the items, and correspondingly attend only to the abstract commonalities of items in sequential presentations.
Our model explains this effect as the result of general memory and attention mechanisms that have been shown to play a role in word learning more widely (cf.~\citeA{nematzadeh.etal.2012.cmcl,nematzadeh.etal.2013.cogsci}).
Interestingly, attention in our model is a function of the token frequency of word--feature co-occurrences (as opposed to a fixed parameter) and is therefore a response to the statistics of the data, as are other components of our word generalization formulation.
All this further supports that attention, memory and statistical learning interact to produce the suspicious coincidence effect and its reversal across presentations.

\section{Conclusions and Future Work}

Novel word generalization -- understanding how a word maps to the appropriate level of a taxonomic hierarchy -- is an important aspect of novel word learning, but one that has not received much attention in the word-learning community.
We propose a unified model of word learning that accounts for the various observed patterns of novel word generalization -- in particular, the suspicious coincidence effect \cite{xu.tenenbaum.2007.psyrev} and its reversal under differing presentation conditions \cite{spencer.etal.2011}.
We extend the model of \citeA{nematzadeh.etal.2015.emnlp} with a novel integration of the general cognitive mechanisms of memory and attention, and show that our model's success is a result of the interaction of forgetting and attention to novelty of word--feature co-occurrences.
Our approach builds on the earlier \NGS model in highlighting the importance of type and token frequency patterns in the input to capturing interesting generalization effects, but here these patterns are manifest in our formulation of memory and attention mechanisms.

In incorporating these cognitive processes into our model, we drew on the approach of \citeA{nematzadeh.etal.2012.cmcl}, whose model had been shown to account for various spacing effects in word learning \cite<see also>{nematzadeh.etal.2013.cogsci}.
Much further work is needed to explore whether our model can explain other such effects.
For example, \citeA{vong.etal.2014} showed that people's categorization of novel object instances depends on the distribution of training examples both that are labelled with a word as well as those that are unlabelled.
Currently, our model only takes into account word--feature co-occurrences, and is therefore insensitive to features that occur without a word label.
We will need to consider how to integrate learning from unlabelled data in order to better model how statistical word learning interacts with object categorization, as it does in people.

\setlength{\bibleftmargin}{.125in}
\setlength{\bibindent}{-\bibleftmargin}
\bibliographystyle{apacite}
\bibliography{nematzadeh}

\end{document}